\documentclass[11pt,a4paper]{article}
\usepackage{naaclhlt2019}  % [nohyperref]
\usepackage{times}
\usepackage{latexsym}
\usepackage{natbib}	  % for citep, etc.
\usepackage{CJKutf8}  % for Japanese text
\usepackage{booktabs}
\usepackage{amsmath,amssymb}
\usepackage{multirow}
\usepackage{xcolor}
\usepackage{url}
\usepackage{graphicx}

\newcommand{\undset}[2]{\underset{#1}{#2\strut}}
\aclfinalcopy % Uncomment this line for the final submission
\def\aclpaperid{2189} %  Enter the acl Paper ID here

\title{Lost in Interpretation:\\Predicting Untranslated Terminology in Simultaneous Interpretation}

\author{
    Nikolai Vogler \and Craig Stewart \and Graham Neubig\\
    Language Technologies Institute \\
    Carnegie Mellon University \\
    {\tt $\{$nikolaiv,cas1,gneubig$\}$@cs.cmu.edu}
}

\date{}

\begin{document}
\maketitle

\begin{abstract}
    Simultaneous interpretation, the translation of speech from one language to another in real-time, is an inherently difficult and strenuous task. 
    One of the greatest challenges faced by interpreters is the accurate translation of difficult terminology like proper names, numbers, or other entities. 
    Intelligent computer-assisted interpreting (CAI) tools that could analyze the spoken word and detect terms likely to be untranslated by an interpreter could reduce translation error and improve interpreter performance. 
    In this paper, we propose a task of predicting which terminology simultaneous interpreters will leave untranslated, and examine methods that perform this task using supervised sequence taggers. 
    We describe a number of task-specific features explicitly designed to indicate when an interpreter may struggle with translating a word. Experimental results on a newly-annotated version of the NAIST Simultaneous Translation Corpus \citep{shimizu2014collection} indicate the promise of our proposed method.\footnote{Code is available at \url{https://github.com/nvog/lost-in-interpretation}. Term annotations for the NAIST Simultaneous Translation Corpus will be provided upon request after confirmation that you have access to the corpus, available at  \url{https://ahcweb01.naist.jp/resource/stc/}.}
\end{abstract}

\section{Introduction}\label{sec:intro}

Simultaneous interpretation (SI) is the act of translating speech in real-time with minimal delay, and is crucial in facilitating international commerce, government meetings, or judicial settings involving non-native language speakers \citep{bendazzoli05epic,hewitt1998court}.
However, SI is a cognitively demanding task that requires both active listening to the speaker and careful monitoring of the interpreter's own output.
Even accomplished interpreters with years of training can struggle with unfamiliar concepts, fast-paced speakers, or memory constraints \citep{lambert1994bridging,liu2004working}.
Human short-term memory is particularly at odds with the simultaneous interpreter as he or she must consistently recall and translate specific terminology uttered by the speaker \citep{lederer1978simultaneous,daro1994verbal}.
Despite psychological findings that rare words have long access times \citep{balota1985locus,jescheniak1994word,griffin1998constraint}, listeners expect interpreters to quickly understand the source words and generate accurate translations.
Therefore, professional simultaneous interpreters often work in pairs \citep{millan2012routledge}; while one interpreter performs, the other notes certain challenging items, such as dates, lists, names, or numbers \citep{jones02conferenceinterpreting}.

\begin{figure*}
\centering
\includegraphics{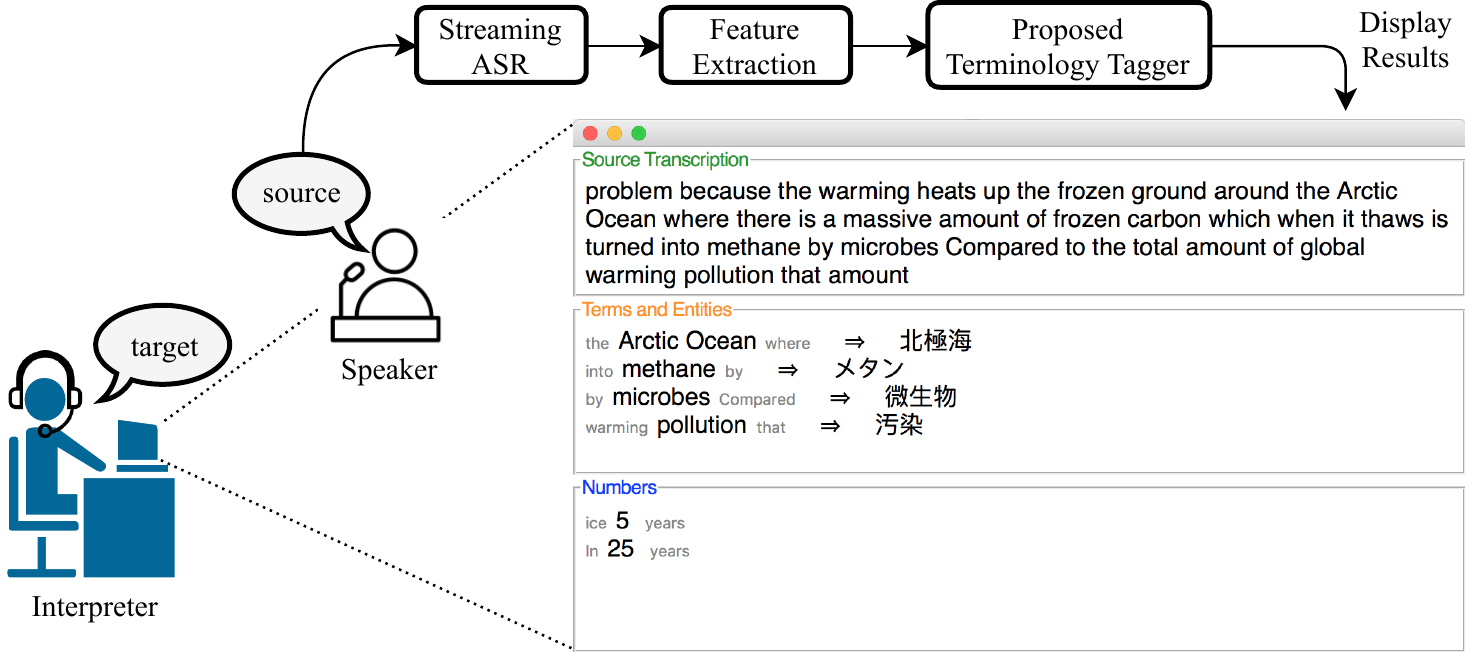}
\caption{
    The simultaneous interpretation process, which could be augmented by our proposed terminology tagger embedded in a computer-assisted interpreting interface on the interpreter's computer.
    In this system, automatic speech recognition transcribes the source speech, from which features are extracted, input into the tagger, and term predictions are displayed on the interface in real-time.
    Finally, machine translations of the terms can be suggested.}
\label{fig:cai}
\end{figure*}

Computers are ideally suited to the task of recalling items given their ability to store large amounts of information, which can be accessed almost instantaneously.
As a result, there has been recent interest in developing computer-assisted interpretation (CAI; \citet{interpretershelp,fantinuoli2016interpretbank,fantinuoli2017speech}) tools that have the ability to display glossary terms mentioned by a speaker, such as names, numbers, and entities, to an interpreter in a real-time setting.
Such systems have the potential to reduce cognitive load on interpreters by allowing them to concentrate on fluent and accurate production of the target message.

These tools rely on automatic speech recognition (ASR) to transcribe the source speech, and display terms occurring in a prepared glossary.
While displaying all terminology in a glossary achieves \textit{high recall} of terms, it suffers from \textit{low precision}.
This could potentially have the unwanted effect of cognitively overwhelming the interpreter with too many term suggestions \cite{stewart2018automatic}.
Thus, an important desideratum of this technology is to only provide terminology assistance when the interpreter requires it.
For instance, an NLP tool that learns to predict only terms an interpreter is likely to miss could be integrated into a CAI system, as suggested in Fig.~\ref{fig:cai}.

In this paper, we introduce the task of predicting the terminology that simultaneous interpreters are likely to leave untranslated using \emph{only} information about the source speech and text.
We approach the task by implementing a supervised, sliding window, SVM-based tagger imbued with delexicalized features designed to capture whether words are likely to be missed by an interpreter.
We additionally contribute new manual annotations for untranslated terminology on a seven talk subset of an existing interpreted TED talk corpus \citep{shimizu2014collection}.
In experiments on the newly-annotated data, we find that intelligent term prediction can increase average precision over the heuristic baseline by up to 30\%.

\section{Untranslated Terminology in SI}\label{sec:taskintro}

Before we describe our supervised model to predict untranslated terminology in SI, we first define the task and describe how to create annotated data for model training.

\subsection{Defining Untranslated Terminology}\label{sec:difficult_terminology}

Formally, we define untranslated terminology with respect to a source sentence $S$, sentence created by a translator $R$, and sentence created by an interpreter $I$. Specifically, we define any consecutive sequence of words $s_{i:j}$, where $0 \leq i \leq N-1$ (inclusive) and $i < j \leq N$ (exclusive), in source sentence $S_{0:N}$ that satisfies the following criteria to be an untranslated term:

\begin{itemize}
\item \textbf{Termhood:} It consists of only numbers or nouns. We specifically focus on numbers or nouns for two reasons: (1) based on the interpretation literature, these categories contain items that are most consistently difficult to recall \citep{jones02conferenceinterpreting,gile2009basic}, and (2) these words tend to have less ambiguity in their translations than other types of words, making it easier to have confidence in the translations proposed to interpreters.

\item \textbf{Relevance:} A translation of $s_{i:j}$, we denote $t$, occurs in a sentence-aligned reference translation $R$ produced by a translator in an offline setting. This indicates that in a time-unconstrained scenario, the term \emph{should} be translated.
\item \textbf{Interpreter Coverage:} It is not translated, literally or non-literally, by the interpreter in interpreter output $I$. This may reasonably allow us to conclude that translation thereof may have presented a challenge, resulting in the content not being conveyed.
\end{itemize}

Importantly, we note that the phrase \textit{untranslated} terminology entails words that are either dropped mistakenly, intentionally due to the interpreter deciding they are unnecessary to carry across the meaning, or mistranslated.
We contrast this with \textit{literal} and \textit{non-literal} term coverage, which encompasses words translated in a verbatim and a paraphrastic way, respectively.

\subsection{Creating Term Annotations}\label{sec:annotation}

To obtain data with labels that satisfy the previous definition of untranslated terminology, we can leverage existing corpora containing sentence-aligned source, translation, and simultaneous interpretation data.
Several of these resources exist, such as the NAIST Simultaneous Translation Corpus (STC) \citep{shimizu2014collection} and the European Parliament Translation and Interpreting Corpus (EPTIC) \citep{bernardini2016eptic}.
Next, we process the source sentences, identifying terms that satisfy the termhood, relevance, and interpreter coverage criteria listed previously.

\begin{itemize}
\item \textbf{Termhood Tests:} To check termhood for each source word in the input, we first part-of-speech (POS) tag the input, then check the tag of the word and discard any that are not nouns or numbers.
\item \textbf{Relevance and Interpreter Coverage Tests:} Next, we need to measure relevancy (whether a corresponding target-language term appears in translated output), and interpreter coverage (whether a corresponding term \emph{does not} appear in interpreted output).
An approximation to this is whether one of the translations listed in a bilingual dictionary appears in the translated or interpreted outputs respectively, and as a first pass we identify all source terms with the corresponding target-language translations.
However, we found that this automatic method did not suffice to identify many terms due to lack of dictionary coverage and also to non-literal translations.
To further improve the accuracy of the annotations, we commissioned human translators to annotate whether a particular source term is translated literally, non-literally, or untranslated by the translator or interpreters (details given in \S\ref{sec:annotation_stc}).
\end{itemize}

Once these inclusion criteria are calculated, we can convert all untranslated terms into an appropriate format conducive to training supervised taggers. In this case, we use an IO tagging scheme \citep{ramshaw1999text} where all words corresponding to untranslated terms are assigned the label I, and all others are assigned a label O, as shown in Fig.~\ref{fig:labeled_sentence}.

\begin{figure}[t]
\centering
\begin{tabular}{@{}c@{\hspace{0.8\tabcolsep}}l@{}}
{\small Src} & \fbox{
	\parbox{.4\textwidth}{
    $\undset{\text{O}}{\text{\small In}}$ 
    $\undset{\text{O}}{\text{\small California}}$, 
    $\undset{\text{O}}{\text{\small there}}$ 
    $\undset{\text{O}}{\text{\small has}}$ 
    $\undset{\text{O}}{\text{\small been}}$ 
    $\undset{\text{O}}{\text{\small a}}$ 
    $\undset{\text{I}}{\lbrack\text{\small 40}\rbrack}$ 
    $\undset{\text{O}}{\text{\small percent}}$ 
    $\undset{\text{O}}{\text{\small decline}}$ 
    $\undset{\text{O}}{\text{\small in}}$ 
    $\undset{\text{O}}{\text{\small the}}$ 
      $\lbrack\undset{\text{I}}{\text{\small Sierra}}$ 
      $\undset{\text{I}}{\text{\small snowpack}}\rbrack$.}} \\

{\small Interp} & \fbox{
  \parbox{.4\textwidth}{
    \begin{CJK}{UTF8}{min}
      $\undset{\text{California}}{\text{\small カリフォルニア で は}}$、 
      $\undset{\text{4}}{\text{\small 4}}$
      $\undset{\text{percent}}{\text{\small パーセント}}$ 
      $\undset{\text{decline}}{\text{\small 少な く な っ て}}$
      {\small しま い ま し た 。}
    \end{CJK}
}}\\
\end{tabular}
\caption{A source sentence and its corresponding interpretation. Untranslated terms are surrounded by brackets and each word in the term is labeled with an I-tag. The interpreter mistakes the term \textit{40} for \textit{4}, and omits \textit{Sierra snowpack}.}
\vspace{-2mm}
\label{fig:labeled_sentence}
\end{figure}

\begin{figure*}
\centering
\includegraphics[trim={2cm 12.58cm 7cm 12cm},clip,width=0.92\textwidth]{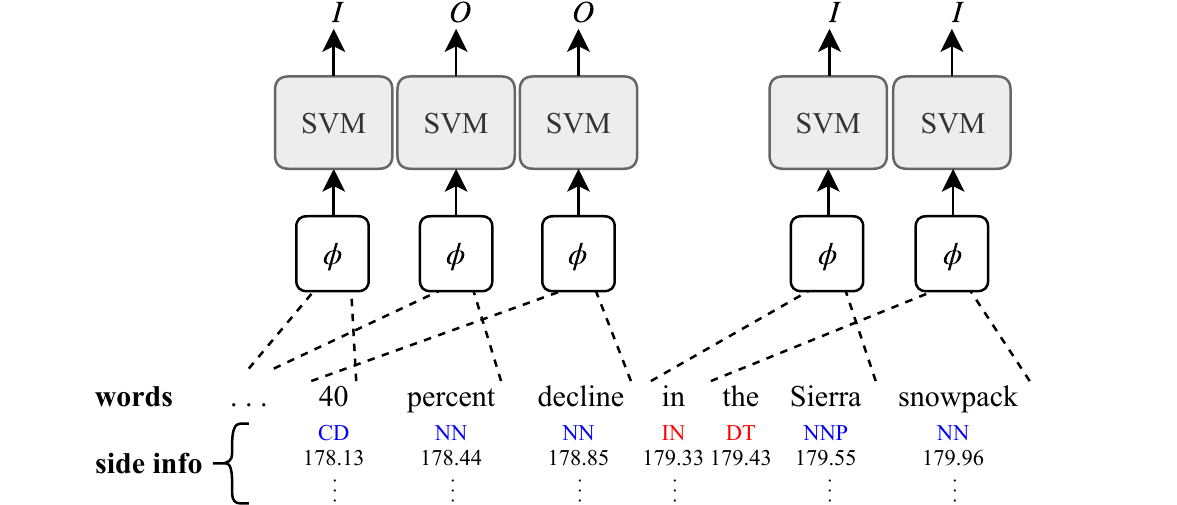}
\caption{
    Our tagging model at prediction time. A sliding window SVM, informed by a task-specific feature function $\phi$ with access to the POS tags, source speech timing (in seconds), and other information, predicts whether or not words matching the termhood constraint (in blue) are likely to be left untranslated in SI.}
\label{fig:model}
\end{figure*}

\section{Predicting Untranslated Terminology}

With supervised training data in hand, we can create a model for predicting untranslated terminology that could potentially be used to provide interpreters with real-time assistance.
In this section, we outline a couple baseline models, and then describe an SVM-based tagging model, which we specifically tailor to untranslated terminology prediction for SI by introducing a number of hand-crafted features.

\subsection{Heuristic Baselines}\label{sec:baselines}
In order to compare with current methods for term suggestion in CAI, such as \citet{fantinuoli2017challenges}, we first introduce a couple of heuristic baselines.
\begin{itemize}
\item \textbf{Select noun/\# POS tag:}
Our first baseline recalls all words that meet the termhood requirement from \S\ref{sec:taskintro}.
Thus, it will achieve perfect recall at the cost of precision, which will equal the percentage of I-tags in the data.
\item \textbf{Optimal frequency threshold:}
To increase precision over this naive baseline, we also experiment with a baseline that has a frequency threshold, and only output words that are rarer than this frequency threshold in a large web corpus, with the motivation that rarer words are more likely to be difficult for translators to recall and be left untranslated.
\end{itemize}

\subsection{SVM-based Tagging Model}
\label{sec:svmtagger}

While these baselines are simple and intuitive, we argue that there are a large number of other features that indicate whether an interpreter is likely to leave a term untranslated. 
We thus define these features, and resort to machine-learned classifiers to integrate them and improve performance.
State-of-the-art sequence tagging models process sequences in both directions prior to making a globally normalized prediction for each item in the sequence \citep{huang2015bidirectional,ma2016end}.
However, the streaming, real-time nature of simultaneous interpretation constrains our model to sequentially process data from left-to-right and make local, monotonic predictions (as noted in \citet{oda14acl,grissom14finalverb}, among others).
Therefore, we use a sliding-window, linear support vector machine (SVM) classifier \citep{cortes1995support,joachims98svm}  that uses only local features of the history to make independent predictions, as depicted in Fig.~\ref{fig:model}.\footnote{We also experimented with a unidirectional LSTM tagger \citep{hochreiter97lstm,graves2012sequence}, but found it ineffective on our small amount of annotated data.}
Formally, given a sequence of source words with their side information (such as timings or POS tags) $S = s_{0:N}$, we slide a window $W$ of size $k$ incrementally across $S$, extracting features $\phi(s_{i-k+1:i+1})$ from $s_i$ and its $k-1$ predecessors.

Since our definition of terminology only allows for nouns and numbers, we restrict prediction to words of the corresponding POS tags $Q = \{$CD, NN, NNS, NNP, NNPS$\}$ using the Stanford POS tagger \citep{toutanova2003feature}.
That is, we assign a POS tag $p_i$ to each word from $s_i$ and only extract features/predict using the classifier if $p_i \in Q$; otherwise we always assign the Outside tag.
This disallows words that are of other POS tags from being classified as untranslated terminology and greatly reduces the class imbalance issue when training the classifier.%
\footnote{We note that a streaming POS tagger would have to be used in a real-time setting, as in \cite{oda15acl}.}

\subsection{Task-specific Features}\label{sec:features}

Due to the fact that only a small amount of human-interpreted human-annotated data can be created for this task, it is imperative that we give the model the precise information it needs to generalize well.
To this end, we propose multiple task-specific, non-lexical features to inform the classifier about certain patterns that may indicate terminology likely to be left untranslated.

\begin{itemize}
\item \textbf{Elapsed time:} 
As discussed in \S\ref{sec:intro}, SI is a cognitively demanding task.
Interpreters often work in pairs and usually swap between active duty and notetaking roles every 15-20 minutes \citep{lambert1994bridging}.
Towards the end of talks or long sentences, an interpreter may become fatigued or face working memory issues---especially if working alone. 
Thus, we monitor the number of minutes elapsed in the talk and the index of the word in the talk/current sentence to inform the classifier.
\item \textbf{Word timing:} 
We intuit that a presenter's quick speaking rate can cause the simultaneous interpreter to potentially drop some terminology.
We obtain word timing information from the source speech via forced alignment tools \citep{ochshorn2016gentle, povey11kaldi}.
The feature function extracts both the number of words in the past $m$ seconds and the time deltas between the current word and previous words in the window.
\item \textbf{Word frequency:} 
We anticipate that interpreters often leave rarer source words untranslated because they are probably more difficult to recall from memory.
On the other hand, we would expect loan words, words adopted from a foreign language with little or no modification, to be easier to recognize and translate for an interpreter.
We extract the binned unigram frequency of the current source word from the large monolingual Google Web 1T Ngrams corpus \citep{brants2006web}. 
We define a loan word as an English word with a Katakana translation in the bilingual dictionaries \citep{eijiro,breen2004jmdict}.
\item \textbf{Word characteristics and syntactic features:} We extract the number of characters and number of syllables in the word, as determined by lookup in the CMU Pronunciation dictionary \citep{weide1998cmu}. 
Numbers are converted to their word form prior to dictionary lookup.
Generally, we expect longer words, both by character and syllable count, to represent more technical or marked vocabulary, which may be challenging to translate.
Additionally, we syntactically inform the model with POS tags and regular expression patterns for numerals.
\end{itemize}

These features are extracted via sliding a window over the sentence, as displayed in Fig.~\ref{fig:model} and discussed in \S\ref{sec:svmtagger}.
Thus, we also utilize previous information from the window when predicting for the current word.
This previous information includes past predictions, word characteristics and syntax, and source speech timing.

\section{Experimental Annotation and Analysis}\label{sec:annotation_stc}

In this section, we detail our application of the term annotation procedure in \S\ref{sec:taskintro} to an SI corpus and analyze our results.

\subsection{Annotation of NAIST STC}
For SI data, we use a seven-talk, manually-aligned subset of the English-to-Japanese NAIST STC \citep{shimizu2014collection}, which consists of source subtitle transcripts, En$\rightarrow$Ja offline translations, and interpretations of English TED talk videos from professional simultaneous interpreters with 1, 4, and 15 years of experience, who are dubbed B-rank, A-rank, and S-rank\footnote{\{B, A, S\}-rank is the Japanese equivalent to \{C, B, A\}-rank on the international scale.}.
TED talks offer a unique and challenging format for simultaneous interpreters because the speakers typically talk in-depth about a single topic, and such there are many new terms that are difficult for an interpreter to process consistently and reliably.
The prevalence of this difficult terminology presents an interesting testbed for our proposed method.

First, we use the Stanford POS Tagger \citep{toutanova2003feature} on the source subtitle transcripts to identify word chunks with a POS tag in $\{$CD, NN, NNS, NNP, NNPS$\}$, discarding words with other tags.
After performing word segmentation on the Japanese data using KyTea \citep{neubig2011kytea}, we automatically detect for translation coverage between the source subtitles, SI, and translator transcripts with a string-matching program, according to the relevance and coverage tests from \S\ref{sec:taskintro}.
The En$\leftrightarrow$Ja \textsc{Eijiro} (2.1m entries) \citep{eijiro} and \textsc{Edict} (393k entries) \citep{breen2004jmdict} bilingual dictionaries are combined to provide term translations.
Additionally, we construct individual dictionaries for each TED talk with key acronyms, proper names, and other exclusive terms (e.g., \textit{UNESCO}, \textit{CO2}, \textit{conflict-free}, \textit{Pareto-improving}) to increase this automatic coverage.
Nouns are lemmatized prior to lookup in the bilingual dictionary, and we discard any remaining closed-class function words.

While this automatic process is satisfactory for identifying if a translated term occurs in the translator's or interpreters' transcripts (relevancy), it is inadequate for verifying the terms that occur in the translator's transcript, but \textit{not} the interpreters' outputs (interpreter coverage).
Therefore, we commissioned seven professional translators to review and annotate those source terms that could not be marked as translated by the automatic process as either \textit{translated}, \textit{untranslated}, or \textit{non-literally translated} in each target sentence.
Lastly, we add I-tags to each word in the untranslated terms and O-tags to the words in literally and non-literally translated terms.

\subsection{Annotation Analysis}

\begin{table}[t]
\centering
\begin{tabular}{c c c c c c c c}
\toprule
    & \multicolumn{2}{c}{\textbf{trans.}} & \multicolumn{2}{c}{\textbf{non-lit.}} & \multicolumn{2}{c}{\textbf{raw untrans.}}\\\cmidrule(lr){2-3}\cmidrule(lr){4-5}\cmidrule(lr){6-7}
\textbf{T/I} & \textbf{\#} & \textbf{\%} & \textbf{\#} & \textbf{\%} & \textbf{\#} & \textbf{\%} \\
\midrule
T   	& 2,213 & 80 & 158 & 6 & 401   & 14 \\ 
B 	    & 1,134 & 41 & 92  & 3 & 1,546 & 56 \\
A	    & 1,151 & 42 & 114 & 4 & 1,507 & 54\\
S  	    & 1,531 & 55 & 170 & 6 & 1,071 & 39\\
\bottomrule
\end{tabular}
\caption{
    Translated, non-literally translated, and raw untranslated term annotations obtained in the annotation process using the NAIST STC for (T)ranslator, and \{B,A,S\}-rank SI. Note that these \textit{raw} untranslated term figures are directly from the annotation process, prior to filtering based off of the term relevancy constraint from \S\ref{sec:taskintro}.
}
\label{tab:annotation1}
\end{table}

Table \ref{tab:annotation1} displays the term coverage annotation statistics for the translators and interpreters.
Since translators performed in an offline setting without time constraints, they were able to translate the largest number of source terms into the target language, with 80\% being literally translated, and 6\% being non-literally translated.
On the other hand, interpreters tend to leave many source terms uncovered in their translations.
The A-rank and B-rank interpreters achieve roughly the same level of term coverage, with the A-rank being only slightly more effective than B-rank at translating terms literally and non-literally.
This is in contrast with \citet{shimizu2014collection}'s automatic analysis of translation quality on a three-talk subset, in which A-rank has slightly higher translation error rate and lower BLEU score \citep{papineni02bleu} than the B-rank interpreter.
The most experienced S-rank interpreter leaves 17\% fewer terms than B-rank uncovered in the translations.
More interestingly, the number of non-literally translated terms also correlates with experience-level.
In fact, the S-rank interpreter actually exceeds the translator in the number of non-literal translations produced.
Non-literal translations can occur when the interpreter fully comprehended the source expression, but chose to generate it in a way that better fit the translation in terms of fluency.

In Table \ref{tab:annotation2}, we show the number of terms left untranslated by each interpreter rank after processing our annotations for the relevancy constraint of \S\ref{sec:taskintro}. 
Since the number of per-word I-tags is only slightly higher than the number of untranslated terms, most such terms consist of only a single word of about 6.5 average characters for all ranks.
Capitalized terms (i.e., named entities/locations) constitute about 14\% of B-rank, 13\% of A-rank, and 15\% of S-rank terms.
Numbers represent about 5\% of untranslated terms for each rank.

\begin{table}[t]
\centering
\begin{tabular}{l r r r}
\toprule
~           & ~                 & \multicolumn{2}{c}{\textbf{\% I-tag of}} \\\cmidrule(lr){3-4}
\textbf{SI} 			& \textbf{\# untrans. terms} 	    & \textbf{all} & \textbf{noun/\#} \\
\midrule
B-rank  	& 1,256             & 10.8     & 45.4\\
A-rank	    & 1,206             & 10.4     & 43.6\\
S-rank	    & 812               & 7.0      & 29.6\\
\bottomrule
\end{tabular}
\caption{
    Final untranslated term count and number of I-tags after filtering based off of the \textit{relevancy} constraint (\S\ref{sec:taskintro}). That is, only the raw untranslated source terms that appear in the translator's transcript are truly considered untranslated.
}
\label{tab:annotation2}
\end{table}

\begin{figure}[t]
\centering
\includegraphics[trim={2.25cm 7.7cm 0cm 8cm},clip,width=.49\textwidth]{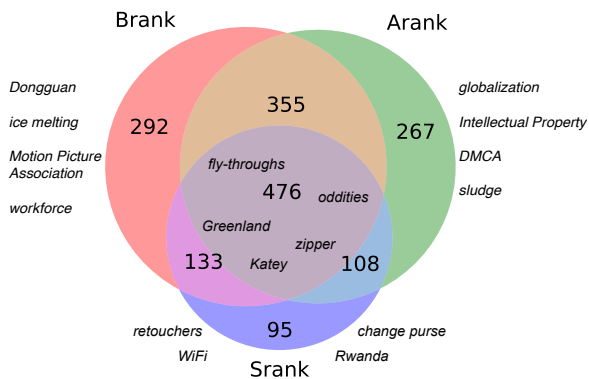}
\caption{Untranslated term overlap between interpreters.}
\label{fig:overlap}
\end{figure}

The untranslated term overlap between interpreters is visualized in Fig.~\ref{fig:overlap}.
Most difficult terms are shared amongst interpreter ranks as only 23.2\% (B), 22.1\% (A), and 11.7\% (S) of terms are unique for each interpreter.
We show a sampling of some unique noun terms on the outside of the Venn diagram, along with the untranslated terms shared among all ranks in the center.
Among these unique terms, capitalized terms make up 19\% of B-rank/S-rank, but only 13\% of A-rank.
7.4\% of S-rank's unique terms are numbers compared with about 5\% for the other two ranks.

\section{Term Prediction Experiments}

\subsection{Experimental Setting}
We design our experiments to evaluate both the effectiveness of a system to predict untranslated terminology in simultaneous interpretation and the usefulness of our features given the small amount of aligned and labeled training data we possess.

We perform leave-one-out cross-validation using five of the seven TED talks as the training set, one as the development set, and one as the test set.
Hyperparameters (SVM's penalty term, the number of bins for the word frequency feature=9, and sliding window size=8) are tuned on the dev. fold and the best model, determined by average precision score, is used for the test fold predictions.
Both training and predictions are performed on a sentence-level.
During training, we weight the two classes inversely proportional to their frequencies in the training data to ensure that the majority O-tag does not dominate the I-tag.

\begin{table}[t]
  \centering
  \begin{tabular}{l c c c}
    \toprule
    ~                           & \multicolumn{3}{c}{\textbf{AP}} \\\cmidrule(lr){2-4}
    \textbf{Method}                           & \textbf{B} & \textbf{A} & \textbf{S}\\
    \midrule
    Select noun/\# POS tag      & 45.4 & 43.6 & 29.6 \\
    Optimal freq threshold      & 49.7 & 48.1 & 32.9 \\
    \midrule
    SVM (all features)          & 58.9 & 53.5 & 39.1 \\
    $-$ elapsed time		    & 58.8 & 53.0 & 38.8 \\
    $-$ word timing             & 58.2 & 53.2 & 38.5 \\
    $-$ word freq 	            & \textbf{59.4} & 52.5 & 39.1 \\
	$-$ characteristic/syntax   & 59.3 & \textbf{55.1} & \textbf{42.5} \\
    \bottomrule
  \end{tabular}
  \caption{
      Average precision score cross-validation results with feature ablation for the untranslated term class on test data. 
      Optimal word frequency threshold is determined on dev set of each fold.
      Evaluation performed on a word-level. 
      Highest numbers per column are bolded. 
      Each setting is statistically significant at $p < 0.05$ by paired bootstrap \citep{koehn04sigtest}.
  }
  \label{tab:results}
\end{table}

\subsection{Results and Analysis}\label{sec:results}

\begin{table}[t]
	\centering
    \begin{tabular}{l l}
    \toprule
            {Select POS} & \parbox{4.7cm}{
            in the last
        	\textcolor{red}{5}
        	\textcolor{red}{years}
        	we
        	've
        	added
        	\textcolor{red}{70000000}
        	\textcolor{red}{tons}
        	of
        	\textcolor{red}{co2}
        	every
            \textcolor{blue}{24}
        	\textcolor{blue}{hours}
        	\textcolor{blue}{25000000}
        	\textcolor{red}{tons}
        	every
        	\textcolor{blue}{day}
        	to
        	the
        	\textcolor{blue}{oceans}
        } \\
    \midrule
        {Optimal freq} & \parbox{4.7cm}{
            in the last
        	5
            years
        	we
        	've
        	added
        	\textcolor{red}{70000000}
        	\textcolor{red}{tons}
        	of
        	\textcolor{red}{co2}
        	every
        	\textcolor{orange}{24}
        	\textcolor{orange}{hours}
        	\textcolor{blue}{25000000}
        	\textcolor{red}{tons}
        	every
        	\textcolor{orange}{day}
        	to
        	the
        	\textcolor{blue}{oceans}
        } \\
    \midrule
        {SVM} & \parbox{4.7cm}{
            in the last
        	5
        	years
        	we
        	've
        	added
        	70000000
        	tons
        	of
        	co2
        	every
        	\textcolor{orange}{24}
        	\textcolor{blue}{hours}
        	\textcolor{blue}{25000000}
        	\textcolor{red}{tons}
        	every
        	\textcolor{orange}{day}
        	to
        	the
        	\textcolor{blue}{oceans}
        } \\
    \bottomrule
    \end{tabular}
    \caption{B-rank output from our model contrasted with baselines. Type I errors are in \textcolor{red}{red}, type II errors in \textcolor{orange}{orange}, and correctly tagged untranslated terminology in \textcolor{blue}{blue}.}
    \label{tab:example}
\end{table}

\begin{figure*}[t]
    \centering
    \begin{frame}
      \hfil\hfil\includegraphics[width=7cm]{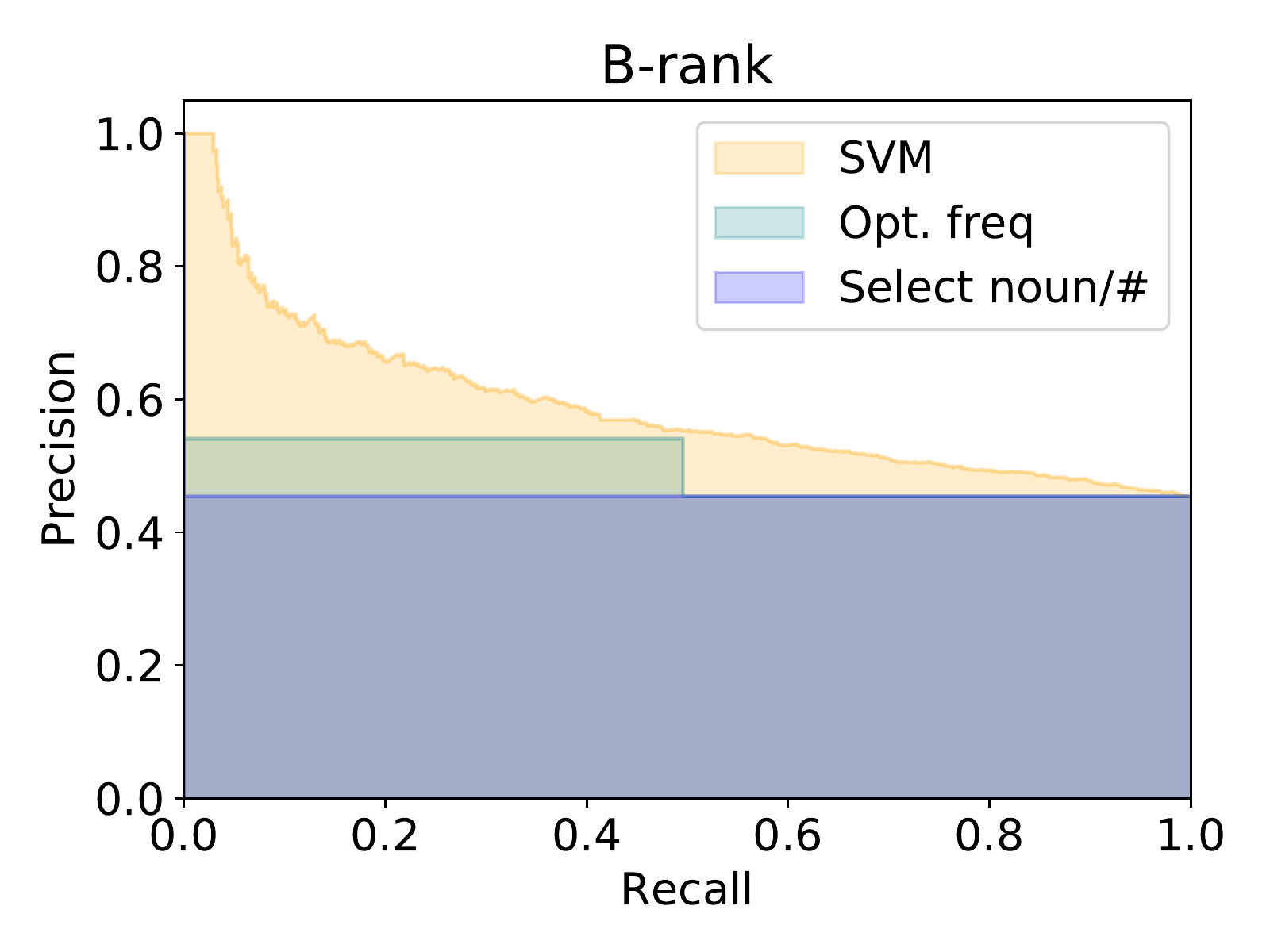}\newline
      \vfil
      \hfil\hfil\includegraphics[width=7cm]{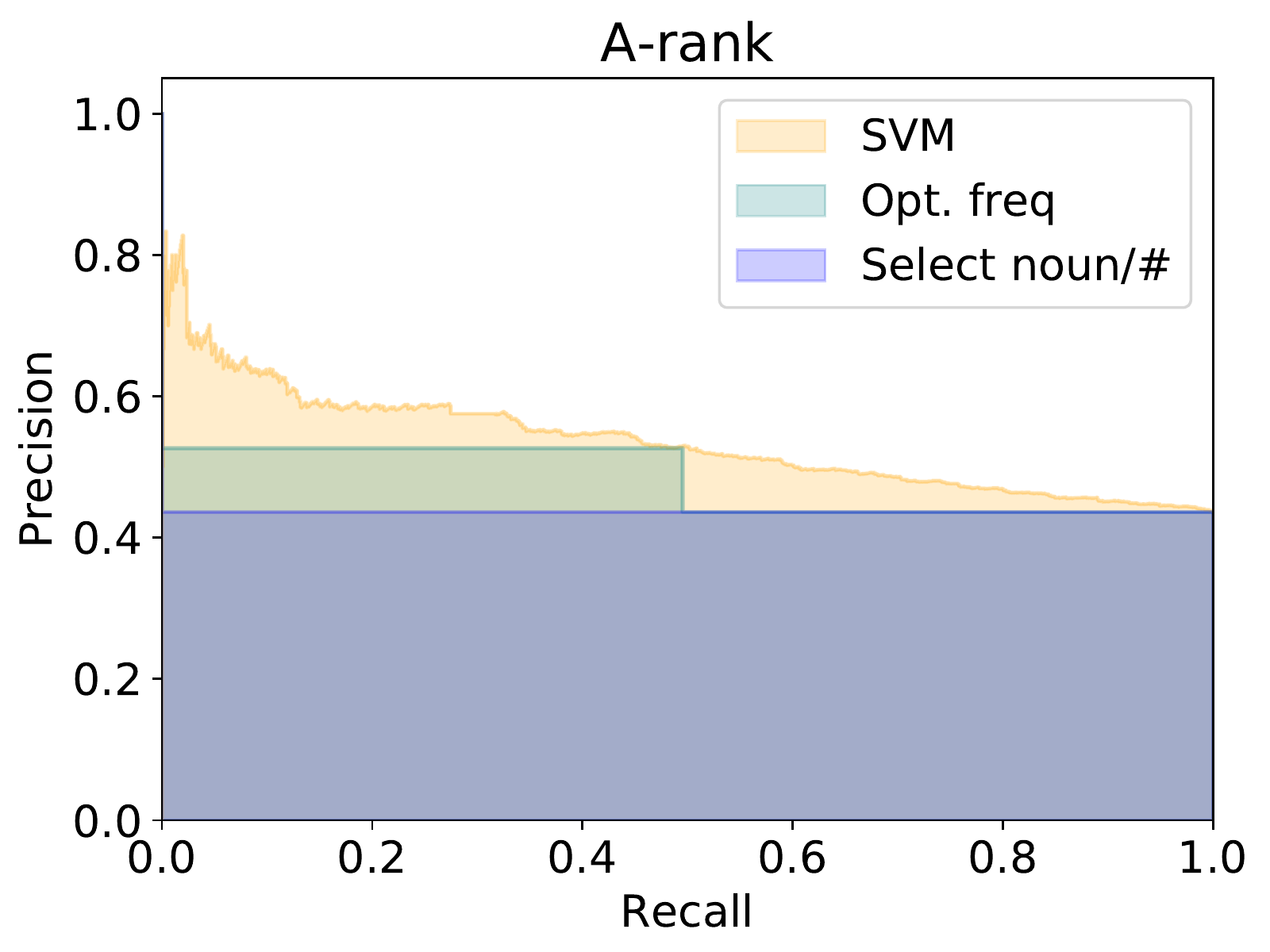}\hfil\hfil
        \includegraphics[width=7cm]{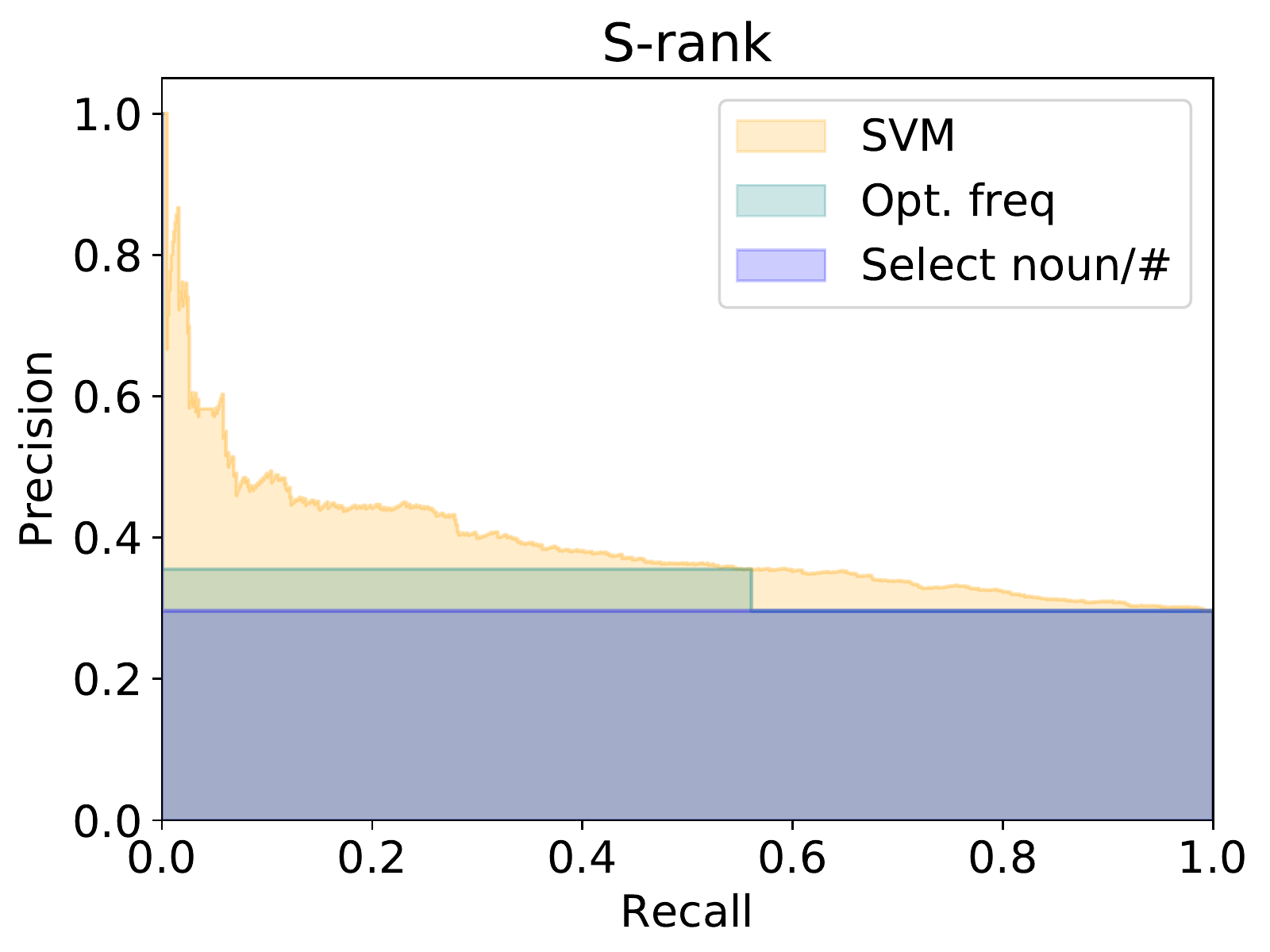}\newline
    \end{frame}
    \caption{Precision-recall curves for each interpreter rank.}
    \label{fig:prcurve}
\end{figure*}

Since we are ultimately interested in the precision and recall trade-off among the methods, we evaluate our results using precision-recall curves in Fig. \ref{fig:prcurve} and the average precision (AP) scores in Table \ref{tab:results}.
AP\footnote{We compute AP using the scikit-learn implementation \citep{scikit-learn}.} summarizes the precision-recall curve by calculating the weighted mean of the precisions at each threshold, where the weights are equal to the increase in recall from the previous threshold.
If the method is embedded in a CAI system, then the user could theoretically adjust the precision-recall threshold to balance helpful term suggestions with cognitive load.

Overall, we tend to see that all methods perform best when tested on data from the B-rank interpreter, and observe a decline in performance across all methods with an increase in interpreter experience.
We believe that this is due to a decrease in the number of untranslated terminology as experience increases (i.e., class imbalance) coupled with the difficulty of predicting such exclusive word occurrences from only source speech and textual cues.
Ablation results in Table \ref{tab:results} show that not all of the features are able to improve classifier performance for all interpreters.
While the elapsed time and word timing features tend to cause a degradation in performance when removed, ablating the word frequency and characteristic/syntax features can actually improve average precision score.
Word frequency, which is a recall-based feature, seems to be more helpful for B- and S-rank interpreters because it is challenging to recall the smaller number of untranslated terms from the data.
Although the characteristic/syntax features are also recall-based, we see a decline in performance for them across all interpreter ranks because they are simply too noisy.
When ablating the uninformative features for each rank, the SVM is able to increase AP vs. the optimal word frequency baseline by about 20\%, 15\%, and 30\% for the B, A, and S-rank interpreters, respectively.

In Table \ref{tab:example}, we show an example taken from the first test fold with results from each of the three methods.
The SVM's increased precision is able to greatly reduce the number of false positives, which we argue could overwhelm the interpreter if left unfiltered and shown on a CAI system.
Nevertheless, one of the most apparent false positive errors that still occurs with our method is on units following numbers, such as the word \textit{tons} in the example.
Also, because our model prioritizes avoiding this type I error, it is more susceptible to type II errors, such as ignoring untranslated terms \textit{24} and \textit{day}.
A user study with our method embedded in a CAI would reveal the true costs of these different errors, but we leave this to future work.

\section{Conclusion and Future Work}
In this paper, we introduce the task of automatically predicting terminology likely to be left untranslated in simultaneous interpretation, create annotated data from the NAIST ST corpus, and propose a sliding window, SVM-based tagger with task-specific features to perform predictions.

We plan to assess the effectiveness of our approach in the near future by integrating it in a heads-up display CAI system and performing a user study. 
In this study, we hope to discover the ideal precision and recall tradeoff point regarding cognitive load in CAI terminology assistance and use this feedback to adjust the model.

Other future work could examine the effectiveness of the approach in the opposite direction (Japanese to English) or on other language pairs.
Additionally, speech features could be extracted from the source or interpreter audio to reduce the dependence on a strong ASR system.

\section{Acknowledgements}
This material is based upon work supported by the National Science Foundation Graduate Research Fellowship under Grant No. DGE1745016 and National Science Foundation EAGER under Grant No. 1748642.
We would like to thank Jordan Boyd-Graber, Hal Daum\'{e} III and Leah Findlater for helpful discussions, Arnav Kumar for assistance with the term annotation interface, and 
the anonymous reviewers for their useful feedback.

\end{document}